\documentclass[english,letterpaper, 10 pt, journal, twoside]{IEEEtran}
\usepackage[T1]{fontenc}
\usepackage[utf8]{inputenc}
\usepackage{color}
\usepackage{array}
\usepackage{refstyle}
\usepackage{url}
\usepackage{multirow}
\usepackage{amsmath}
\usepackage{amssymb}
\usepackage{graphicx}
\usepackage{rotating}

\makeatletter

\AtBeginDocument{\providecommand\Figref[1]{\ref{Fig:#1}}}
\AtBeginDocument{\providecommand\Subsecref[1]{\ref{Subsec:#1}}}
\AtBeginDocument{\providecommand\Tabref[1]{\ref{Tab:#1}}}
\providecommand{\tabularnewline}{\\}
\RS@ifundefined{subsecref}
  {\newref{subsec}{name = \RSsectxt}}
  {}
\RS@ifundefined{thmref}
  {\def\RSthmtxt{theorem~}\newref{thm}{name = \RSthmtxt}}
  {}
\RS@ifundefined{lemref}
  {\def\RSlemtxt{lemma~}\newref{lem}{name = \RSlemtxt}}
  {}

\usepackage{xcolor}         
\usepackage{pifont}         
\usepackage{cite}
\usepackage[caption=false]{subfig}
\usepackage{colortbl}
\usepackage{booktabs}
\usepackage{graphicx}
\usepackage{caption}

\IEEEoverridecommandlockouts                              

\definecolor{alpha}{HTML}{FFCE19}
\definecolor{bob}{HTML}{FF00FF}
\definecolor{carol}{HTML}{19FFDC}

\let\NAT@parse\undefined
\usepackage[pagebackref,breaklinks,colorlinks]{hyperref}
\makeatletter

\@ifundefined{showcaptionsetup}{}{%
 \PassOptionsToPackage{caption=false}{subfig}}
\usepackage{subfig}
\makeatother

\usepackage{babel}
\begin{document}
\title{CaRtGS: Computational Alignment for Real-Time Gaussian Splatting SLAM}
\author{Dapeng Feng, Zhiqiang Chen, Yizhen Yin, \\
Shipeng Zhong, Yuhua Qi, and Hongbo Chen\thanks{Manuscript received: October 6, 2024; Revised January 13, 2025; Accepted
February 19, 2025. This paper was recommended for publication by Editor
Sven Behnke upon evaluation of the Associate Editor and Reviewers'
comments. Corresponding authors: Hongbo Chen and Yuhua Qi.}\thanks{Dapeng Feng, Yizhen Yin, Yuhua Qi, and Hongbo Chen are with Sun Yat-sen
University, Guangzhou, China. (e-mail: \{fengdp5, yinyzh5\}@mail2.sysu.edu.cn,
\{qiyh8, chenhongbo\}@mail.sysu.edu.cn)}\thanks{Zhiqiang Chen is with The University of Hong Kong, Hong Kong SAR,
China. (e-mail: zhqchen@connect.hku.hk)}\thanks{Shipeng Zhong is with WeRide Inc., Guangzhou, China. (e-mail: shipeng.zhong@weride.ai)}\thanks{Digital Object Identifier (DOI): see top of this page.}}
\maketitle
\begin{abstract}
Simultaneous Localization and Mapping (SLAM) is pivotal in robotics,
with photorealistic scene reconstruction emerging as a key challenge.
To address this, we introduce Computational Alignment for Real-Time
Gaussian Splatting SLAM (CaRtGS), a novel method enhancing the efficiency
and quality of photorealistic scene reconstruction in real-time environments.
Leveraging 3D Gaussian Splatting (3DGS), CaRtGS achieves superior
rendering quality and processing speed, which is crucial for scene
photorealistic reconstruction. Our approach tackles computational
misalignment in Gaussian Splatting SLAM (GS-SLAM) through an adaptive
strategy that enhances optimization iterations, addresses long-tail
optimization, and refines densification. Experiments on\textcolor{red}{{}
}Replica, TUM-RGBD, and VECtor datasets demonstrate CaRtGS's effectiveness
in achieving high-fidelity rendering with fewer Gaussian primitives.
This work propels SLAM towards real-time, photorealistic dense rendering,
significantly advancing photorealistic scene representation. For the
benefit of the research community, we release the code and accompanying
videos on our project website: \url{https://dapengfeng.github.io/cartgs}.
\end{abstract}


\section{Introduction}

\IEEEPARstart{S}{imultaneous} Localization and Mapping (SLAM) is
a cornerstone of robotics and has been a subject of extensive research
over the past few decades \cite{teed2021droid,segal2009generalized,campos2021orb,zhong2024colrio,feng2024s3e}.
The rapid evolution of applications such as autonomous driving, virtual
and augmented reality, and embodied intelligence has introduced new
challenges that extend beyond the traditional scope of real-time tracking
and mapping. Among these challenges is the need for photorealistic
scene reconstruction, which necessitates precise spatial understanding
coupled with high-fidelity visual representation.

In response to these challenges, recent research has explored the
use of implicit volumetric scene representations, notably Neural Radiance
Fields (NeRF) \cite{mildenhall2021nerf}. While promising, integrating
NeRF into SLAM systems has encountered several obstacles, including
high computational demands, lengthy optimization times, limited generalizability,
an over-reliance on visual cues, and a susceptibility to catastrophic
forgetting \cite{tosi2024nerfs}.

In a significant breakthrough, a novel explicit scene representation
method utilizing 3D Gaussian Splatting (3DGS) \cite{kerbl20233d}
has emerged as a potent solution. This method not only rivals the
rendering quality of NeRF but also excels in processing speed, offering
an order-of-magnitude improvement in both rendering and optimization
tasks.

The advantages of this representation make it a strong candidate for
incorporation into online SLAM systems that require real-time performance.
It has the potential to transform the field by enabling photorealistic
dense SLAM, thereby expanding the horizons of scene understanding
and representation in dynamic environments.

However, existing Gaussian Splatting SLAM (GS-SLAM) methods \cite{matsuki2024gaussian,keetha2024splatam,yugay2023gaussian,zhu2024loopsplat,sandstrom2024splat,sarikamis2024ig,peng2024rtg,ha2024rgbd,huang2024photo}
struggle to achieve superior rendering performance under real-time
constraints when dealing with a limited number of Gaussian primitives.
These issues stem from the misalignment between the computational
demands of the algorithm and the available processing resources, which
can lead to insufficient optimization and optimization processes.
Addressing these challenges is crucial for enhancing the performance
and applicability of GS-SLAM in real-time environments.

In this paper, we scrutinize the computational misalignment phenomenon
and propose the \textbf{C}omputational \textbf{A}lignment for \textbf{R}eal-\textbf{T}ime
\textbf{G}aussian \textbf{S}platting \textbf{S}LAM (CaRtGS) to address
these challenges. Our approach aims to optimize the computational
efficiency of GS-SLAM, ensuring that it can meet the demands of real-time
applications while achieving high rendering quality with fewer Gaussian
primitives.

Our contributions are listed as follows:
\begin{itemize}
\item We provide an analysis of the computational misalignment phenomenon
present in GS-SLAM.
\item We introduce an adaptive computational alignment strategy that effectively
tackles insufficient optimization, long-tail optimization, and weak-constrained
densification, achieving high-fidelity rendering with fewer Gaussian
primitives under real-time constraints.
\item We conduct comprehensive experiments and ablation studies to demonstrate
the effectiveness of our proposed method on three popular datasets
with three distinct camera types.
\end{itemize}

\section{Related Works}

GS-SLAM leverages the benefits of 3DGS \cite{kerbl20233d} to achieve
enhanced performance in terms of rendering speed and photorealism.
In this section, we conduct a concise review of both 3D Gaussian Splatting
and Gaussian Splatting SLAM.

\subsection{3D Gaussian Splatting}

3DGS \cite{kerbl20233d} is a cutting-edge real-time photorealistic
rendering technology that employs differentiable rasterization, eschewing
traditional volume rendering methods. This groundbreaking method represents
the scene as explicit Gaussian primitivies and enables highly efficient
rendering, achieving a remarkable $1080\text{p}$ resolution at $130$
frames per second (FPS) on contemporary GPUs, and has substantially
spurred research advancements.

In response to the burgeoning interest in 3DGS, a variety of extensions
have been developed with alacrity. Accelerating the acquisition of
3DGS scene representations is a key area of focus, with various strategies
being explored. One prominent research direction is the reduction
of Gaussians through the refinement of densification heuristics \cite{kheradmand20243d,lu2024scaffold,mallick2024taming}.
Moreover, optimizing runtime performance has become a priority, with
several initiatives concentrating on enhancing the differentiable
rasterizer and optimizer implementations \cite{mallick2024taming,durvasula2023distwar,hollein20243dgs,feng2024flashgs}. 

Motivated by these advancements, our work addresses the challenge
of insufficient optimization in photorealistic rendering within real-time
SLAM by utilizing splat-wise backpropagation \cite{mallick2024taming}.
In parallel, recent methodologies have concentrated on sparse-view
reconstruction and have sought to compact the scene representation.
This is achieved by training a neural network to serve as a data-driven
prior, which is capable of directly outputting Gaussians in a single
forward pass \cite{fan2024instantsplat,niedermayr2024compressed,morgenstern2024compact,wang2024end}.
In contrast, our research zeroes in on real-time dense-view and per-scene
visual SLAM. This targeted focus demands an incremental photorealistic
rendering output that is tailored to the unique characteristics of
each scene.

\begin{figure}
\begin{centering}
\includegraphics[width=1\linewidth]{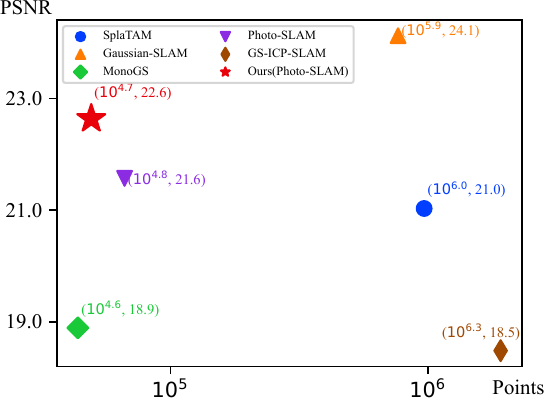}
\par\end{centering}
\caption{\label{fig:Overview-of-GS-SLAM}\textbf{Performance on TUM-RGBD.}
We provide a comparison of most of the available open-source GS-SLAM
methods.}

\vspace{-6mm}
\end{figure}

\subsection{Gaussian Splatting SLAM}

3DGS \cite{kerbl20233d} has also quickly gained attention in the
SLAM literature, owing to its rapid rendering capabilities and explicit
scene representation. MonoGS \cite{matsuki2024gaussian} and SplaTAM
\cite{keetha2024splatam} are seminal contributions to the coupled
GS-SLAM algorithms, pioneering a methodology that simultaneously refines
Gaussian primitives and camera pose estimates through gradient backpropagation.
Gaussian-SLAM \cite{yugay2023gaussian} introduces the concept of
sub-maps to address the issue of catastrophic forgetting. Furthermore,
LoopSplat \cite{zhu2024loopsplat}, which extends the work of Gaussian-SLAM
\cite{yugay2023gaussian}, employs a Gaussian splat-based registration
for loop closure to enhance pose estimation accuracy. However, the
reliance on the intensive computations of 3DGS \cite{kerbl20233d}
for estimating the camera pose of each frame presents challenges for
these methods in achieving real-time performance. 

To overcome this, decoupled GS-SLAM methods have been proposed \cite{sandstrom2024splat,sarikamis2024ig,peng2024rtg,ha2024rgbd,huang2024photo}.
Splat-SLAM \cite{sandstrom2024splat} and IG-SLAM \cite{sarikamis2024ig}
utilize pre-trained dense bundle adjustment \cite{teed2021droid}
for camera pose tracking and proxy depth maps for map optimization.
RTG-SLAM \cite{peng2024rtg} incorporates frame-to-model ICP for tracking
and renders depth by focusing on the most prominent opaque Gaussians.
GS-ICP-SLAM \cite{ha2024rgbd} achieve remarkably high speeds (up
to 107 FPS) by leveraging the shared covariances between G-ICP \cite{segal2009generalized}
and 3DGS \cite{kerbl20233d}, with scale alignment of Gaussian primitives.
Photo-SLAM \cite{huang2024photo} employs ORB-SLAM3 \cite{campos2021orb}
for tracking and introduces a coarse-to-fine map optimization for
robust performance.

These methods achieve state-of-the-art PSNR with a large number of
Gaussian primitives, as presented in \Figref{Overview-of-GS-SLAM},
which will limit the application of real-time GS-SLAM in large-scale
scenarios due to increased computational demands. In this paper, we
delve into the limitations of existing GS-SLAM and propose an innovative
computational alignment technique to enhance PSNR while reducing the
number of Gaussian primitives required, all within the constraints
of real-time SLAM operations.

\section{Methods}

In this section, we delve into the photorealistic rendering aspect
of GS-SLAM. Initially, we scrutinize the computational misalignment
phenomenon inherent to GS-SLAM. This misalignment can significantly
impair computational efficiency and hinder the swift convergence of
photorealistic rendering, adversely affecting the performance of real-time
GS-SLAM. To overcome these obstacles, we propose a novel adaptive
computational alignment strategy. This strategy aims to accelerate
the 3DGS process, optimize computational resource allocation, and
efficiently control model complexity, thereby enhancing the overall
effectiveness and practicality of 3DGS in real-time SLAM applications.

\subsection{\label{subsec:Computational-Misalignment}Computational Misalignment}

The computational misalignment encountered in photorealistic rendering
within the context of SLAM can be attributed to three primary aspects:
insufficient optimization, long-tail optimization, and weak-constrained
densification, which reduces rendering quality and increases map size.
These factors significantly hinder the real-time applications of GS-SLAM,
limiting its applicability in resource-constrained devices.

\begin{figure}
\begin{centering}
\includegraphics[width=1\linewidth]{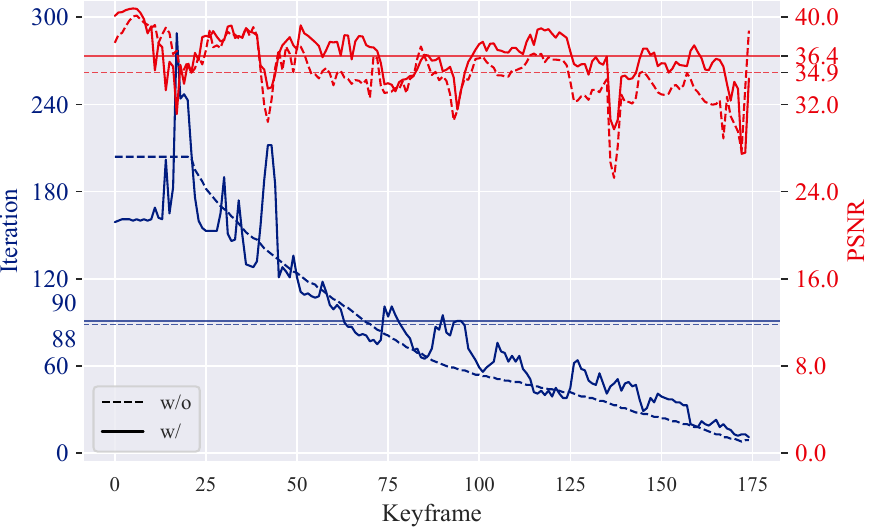}
\par\end{centering}
\caption{{\small{}\label{fig:adaptive-opt}}\textbf{The Effect of Adaptive
Optimization on Replica.} Dashed lines depict performance without
adaptive optimization, while solid lines show results with it. Blue
represents keyframe iterations, and red indicates PSNR. The horizontal
line marks average PSNR and iterations. Our method significantly improves
low-PSNR keyframe processing through enhanced iterative optimization,
as evident from the trend comparison between dashed and solid lines.}

\vspace{-6mm}
\end{figure}

\subsubsection{Insufficient Optimization}

In contrast to typical 3DGS \cite{kerbl20233d}, which is not constrained
by real-time considerations, online rendering within the realm of
SLAM necessitates the concurrent execution of localization, mapping,
and rendering at a speed that is synchronized with the frequency of
incoming sensor data. To achieve this, the majority of current real-time
GS-SLAM methods \cite{peng2024rtg,huang2024photo,ha2024rgbd} rely
on keyframes for both mapping and rendering. However, these methods
typically achieve only a few thousand iterations in rendering optimization
in total, which significantly lags behind the tens of thousands of
iterations achieved by 3DGS \cite{kerbl20233d}. Due to insufficient
optimization, the optimization process has not fully converged, adversely
affecting the quality of online rendering.

Recent observations by several researchers indicate that pixel-wise
backpropagation in 3DGS presents a significant computational challenge
\cite{durvasula2023distwar,mallick2024taming}. This process becomes
a bottleneck due to the contention among multiple GPU threads for
access to shared Gaussian primitives, which necessitates serialized
atomic operations, thereby limiting parallelization efficiency. Unfortunately,
this drawback is integrated into the previous implementations of GS-SLAM
\cite{peng2024rtg,huang2024photo,ha2024rgbd}. In this paper, we utilize
a fast splat-wise backpropagation \cite{mallick2024taming} to reduce
thread contention. This approach not only achieves a $3\times$ increase
in the number of iterations compared to the baseline \cite{huang2024photo},
but also maintains the same runtime. This advancement significantly
mitigates the problem of insufficient optimization, substantially
improving the rendering quality of real-time GS-SLAM.

\begin{figure*}
\begin{centering}
\includegraphics[width=1\linewidth]{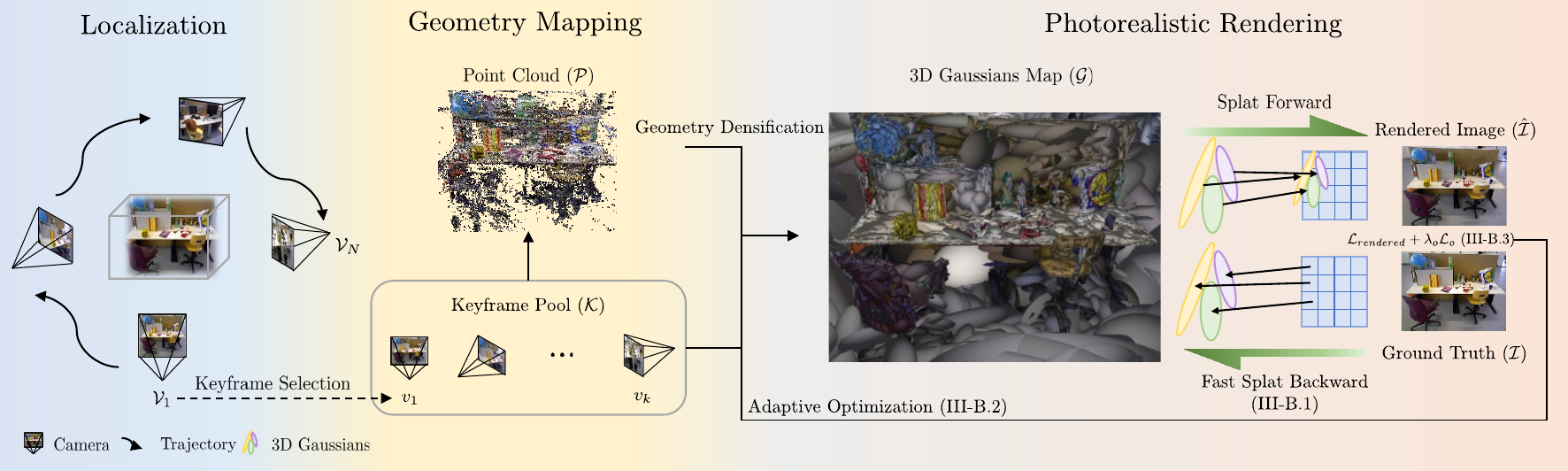}
\par\end{centering}
\caption{{\small{}\label{fig:Overview-of-CaRtGS}}\textbf{\small{}The overview
of CaRtGS. }{\small{}We adopt a real-time cutting-edge SLAM system
as a front-end tracker, severing for localization and geometry mapping.
In the photorealistic rendering back-end, we apply the proposed adaptive
computational alignment strategy to enhance the 3DGS optimization
process, including fast splat backward, adaptive optimization, and
opacity regularization.}}

\vspace{-6mm}
\end{figure*}

\subsubsection{Long-Tail Optimization}

To mitigate the issue of catastrophic forgetting, a common approach
in GS-SLAM is to randomly select a keyframe from the keyframe pool
for periodic reoptimization\cite{huang2024photo,peng2024rtg,ha2024rgbd}.
However, this method can result in suboptimal long-tail optimization,
which overfits the oldest keyframe and underfits the newest one, as
depicted in \Figref{adaptive-opt}. Specifically, the reoptimization
frequency of the earliest keyframes tends to exceed that of the most
recently added ones. This disparity arises because the keyframe pool
is continuously expanded as the camera moves through the environment,
which can result in an uneven distribution of reoptimization efforts
and and a declining trend in the PSNR for newly incoming keyframes.

In this paper, we propose an innovative adaptive optimization strategy
that selects reoptimization keyframes from the pool based on their
optimization loss to counteract long-tail effect. By employing this
approach, we aim to increase the reoptimization frequency of keyframes
with lower PSNR values. This targeted approach has been demonstrated
to significantly enhance the rendering quality, as evidenced by an
improvement from $34.9\,\text{dB}$ to $36.4\,\text{dB}$ in the Replica
Room2 scenario, as depicted in \Figref{adaptive-opt}. By doing so,
our adaptive strategy ensures a more equitable distribution of reoptimization
efforts across the keyframe pool, optimizing each keyframe's contribution
to the system's overall performance. This innovative approach not
only improves the quality of the rendered output but also enhances
the efficiency and effectiveness of the reoptimization process.

\subsubsection{Weak-constrained Densification}

Densification is a critical component of photorealistic rendering
in the context of GS-SLAM, encompassing both geometry densification
and adaptive densification\cite{keetha2024splatam,sarikamis2024ig,yugay2023gaussian,sandstrom2024splat,peng2024rtg,matsuki2024gaussian,huang2024photo,ha2024rgbd,zhu2024loopsplat}.
Geometric densification involves the conversion of a color point cloud
into initialized Gaussian primitives for each newly identified keyframe,
providing a foundational geometric structure for the environment.
Adaptive densification, on the other hand, refines the Gaussian primitives
using operations such as splitting and cloning, which are guided by
gradients and the size of the primitives themselves \cite{kerbl20233d}.
These densifications are solely constrained by a simplistic pruning
strategy that eliminates Gaussian primitives with low opacity. However,
emerging research \cite{niedermayr2024compressed,morgenstern2024compact,wang2024end}
suggests that this approach is insufficient for managing the model's
size within an optimal range. In this paper, we introduce an opacity
regularization loss to encourage the Gaussian primitives to learn
a low opacity, thereby not only facilitating the pruning process to
eliminate less significant primitives but also preserving high-fidelity
rendering.

\subsection{\textcolor{red}{\label{subsec:System-Overview}}System Overview}

As delineated in \Figref{Overview-of-CaRtGS}, we take the modular
designs, which are easy to integrate into existing real-time decoupled
GS-SLAM, e.g., GS-ICP-SLAM\cite{ha2024rgbd} and Photo-SLAM \cite{huang2024photo}.

Given a sequence of observations $\{\mathcal{V}_{1},...,\mathcal{V}_{N}\}$,
we employ a state-of-the-art front-end tracker \cite{segal2009generalized,campos2021orb},
which estimates the 6-DoF pose for each frame and identifies keyframes
$\{v_{1},...,v_{k}\}$ based on criteria related to translation and
rotation. Once a keyframe $v_{i}$ is identified, the frontend tracker
transforms the corresponding observation $\mathcal{V}_{i}$ into the
global coordinate system and integrates it into the global Point Cloud
$\mathcal{P}$.

In the photorealistic rendering phase, we utilize 3DGS \cite{kerbl20233d}
as the backend render. Firstly, we convert $\mathcal{P}$ into a set
of Gaussian primitives $\mathcal{G}$. Each primitive is characterized
by its posistion $\mathbf{p}\in\mathbb{R}^{3}$, orientation represented
as quaternion $\mathbf{q}\in\mathbb{R}^{4}$ , scaling factor $\mathbf{s}\in\mathbb{R}^{3}$,
opacity $\sigma\in\mathbb{R}^{1}$, and spherical harmonic coefficients
$\mathbf{SH}\in\mathbb{R}^{48}$. By employing $\alpha$--blending
rendering \cite{kerbl20233d}, we achieve the high-fidelity rendering
$\hat{\mathcal{I}}$ for a selected keyframe $v_{i}$:

\vspace{-2mm}

\begin{equation}
\hat{\mathcal{I}}=\sum_{k\in\mathcal{G}}c_{k}\alpha_{k}\prod_{j=1}^{k-1}\left(1-\alpha_{k}\right),
\end{equation}

\noindent where $c_{k}$ denotes the color derived from $\mathbf{SH}$,
$\alpha_{k}$ is determined by evaluating a projected 2D Gaussian
multipied with the learned opacity $\sigma_{k}$. To refine the Gaussian
primitives $\mathcal{G}$, we take both $\mathcal{L}_{1}$ and Structural
Similarity Index (SSIM) Loss $\mathcal{L}_{ssim}$ to supervise the
optimization process. These losses are crucial for enhancing the quality
of our photorealistic renderings. Additionally, we incorporate opacity
regularization into our comprehensive loss function to control the
model size, which is detailed in Sec. \Subsecref{Opacity-Regularization}.

\subsection{\label{subsec:Adaptive-Computational-Alignment}Adaptive Computational
Alignment}

To address the computational misalignment of photorealistic rendering
in real-time GS-SLAM, we propose an adaptive computational alignment
strategy termed CaRtGS. Below, we outline the key steps of this strategy
in detail.

\subsubsection{Fast Splat-wise Backpropagation}

\begin{figure*}
\begin{centering}
\par\end{centering}
\begin{minipage}[t]{0.4\linewidth}%
\subfloat[{\small{}\label{fig:splat-wise-backpropagation}}Gradient Backpropagation]{\begin{centering}
\includegraphics[width=1\linewidth]{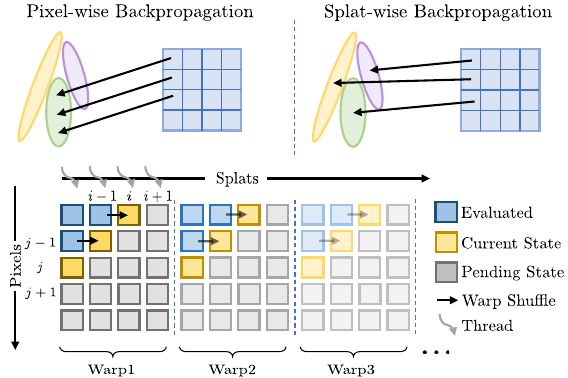}
\par\end{centering}
}%
\end{minipage}\hfill{}%
\begin{minipage}[t]{0.55\linewidth}%
\subfloat[{\small{}\label{fig:total-iteration}Total Iteration on Replica}]{\begin{centering}
\includegraphics[width=1\linewidth]{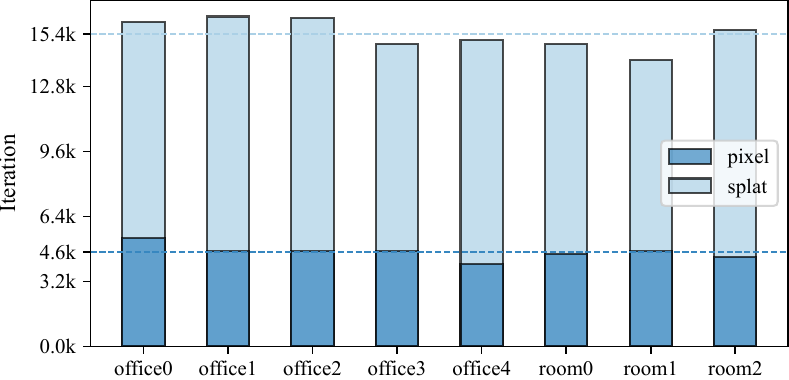}
\par\end{centering}
}%
\end{minipage}

\caption{{\small{}\label{fig:backpropagation}}\textbf{\small{}The Effect of
Different Gradient Backpropagation.} (a) The original 3DGS employs
pixel-wise parallelism for backpropagation, which is prone to frequent
contentions, leading to slower backward passes. We introduce a splat-centric
parallelism, where each thread handles one Gaussian splat at a time,
significantly reducing contention. The gradient computation relies
on a set of per-pixel, per-splat values, effectively traversing a
splat $\Leftrightarrow$ pixel relationship table. During the forward
pass, we save pixel states for every $32^{\text{nd}}$ splat. For
the backward pass, splats are grouped into buckets of $32$, each
processed by a CUDA warp. Warps utilize intra-warp shuffling to efficiently
construct their segment of the state table. {\small{}(b) We provide
a comparison of total iteration on Replica with monocular camera.}}

\vspace{-4mm}
\end{figure*}

In the conventional 3DGS optimization pipeline, the backpropagation
phase is computationally demanding as it entails the propagation of
gradient information from pixels to Gaussian primitives. This process
necessitates the calculation of gradients for each splat-pixel pair
$(i,j)$, followed by an aggregation step. In our notation, $i$ denotes
the index of the $i$-th splat, and $j$ denotes the index of the
$j$-th pixel. To parallelize the execution, we assign thread $i$
to process the $i$-th splat, and thread $j$ to process the $j$-th
pixel. In the forward pass, GPU thread $i+1$ applies the standard
$\alpha$-blending logic to transition from the received state $\mathcal{X}_{i,j}$
to $\mathcal{X}_{i+1,j}$, integrating this updated information into
the gradient computation. In the backward pass, the gradients associated
with the $i$-th splat, denoted as $\nabla\mathcal{X}_{i}$, are accumulated
across the pixels that are influenced by this splat. This process
can be mathematically represented as:

\vspace{-8mm}

\begin{align}
\mathcal{X}_{i+1,j} & =\mathcal{F}(\mathcal{X}_{i,j}),\\
\nabla\mathcal{X}_{i,j} & =\nabla\mathcal{F}\cdot\nabla\mathcal{X}_{i+1,j},\\
\nabla\mathcal{X}_{i} & =\sum_{j}\nabla\mathcal{X}_{i,j},
\end{align}

\noindent where $\mathcal{F}$ presents the $\alpha$-blending function.

Pixel-wise propagation is widely used in GS-SLAM \cite{keetha2024splatam,sarikamis2024ig,yugay2023gaussian,sandstrom2024splat,peng2024rtg,matsuki2024gaussian,huang2024photo,ha2024rgbd,zhu2024loopsplat},
mapping threads to pixels and processing splats in reverse depth order.
Thread $j$ computes partial gradients for the splats in the order
they are blended, updating the cumulative gradient for each splat
through atomic operations. However, this method can lead to contention
among threads for shared memory access, resulting in serialized operations
that impede performance. 

To address this challenge, we utilize a novel parallelization strategy
\cite{mallick2024taming} that shifts the focus from pixel-based to
splat-based processing. This strategy allows each thread to independently
maintain the state of a splat and to efficiently exchange pixel state
information. Thread $i$ can compute the gradient contribution for
the $i$-th splat, requiring the pixel $j$ state after the first
$i$ splats have been blended.

During the forward pass, threads archive transmittance $T$ and accumulated
color $RGB$ for pixels every N splats, preparing for the backward
pass. These stored states include initial conditions $\mathcal{X}_{0,j},\mathcal{X}_{N,j},\cdots\forall j$.
At the commencement of the backward pass, each thread in a tile generates
the pixel state $\mathcal{X}_{i,j}$. Threads then engage in rapid
collaborative sharing to exchange pixel states.

For further details, please refer to \Figref{splat-wise-backpropagation}.
The data presented in \Figref{total-iteration} clearly show that
the splat-wise backpropagation method significantly enhances the total
number of optimization iterations by a factor of $3$, increasing
from an average of $4.6\text{k}$ to $15.4\text{k}$. This improvement
effectively addresses the issue of insufficient optimization compared
to Photo-SLAM \cite{huang2024photo} equipped with pixel-wise propagation.

\subsubsection{Adaptive Optimization}

Although splat-wise propagation achieves sufficient optimization in
total, the long-tail distribution of iterations per keyframe is a
challenge. To address this, we recommend augmenting the splat-wise
approach with an adaptive optimization based on traininig loss $\mathcal{L}$
to ensure a more equitable distribution of iterations across the keyframe
pool $\mathcal{K}$.

Given a keyframe pool $\mathcal{K}_{k}$ containing keyframes $\left\{ v_{1},v_{2},\ldots,v_{k}\right\} $,
we maintain two sets: $\mathcal{R}_{k}=\left\{ r_{1},r_{2},\ldots,r_{k}\right\} $
which tracks the remaining optimization iterations for each keyframe,
and $\mathcal{L}_{k}=\left\{ l_{1},l_{2},\ldots,l_{k}\right\} $ which
records the last optimization loss value for each keyframe. Upon the
detection of a new keyframe $v_{k+1}$, we update our pools as follows:

\vspace{-4mm}

\begin{align}
\mathcal{K}_{k+1} & =\mathcal{K}_{k}\cup\left\{ v_{k+1}\right\} ,\\
\mathcal{R}_{k+1} & =\mathcal{R}_{k}\cup\left\{ r_{k+1}^{0}\right\} ,\\
\mathcal{L}_{k+1} & =\mathcal{L}_{k}\cup\left\{ l_{k+1}\right\} ,
\end{align}

\begin{table*}
\caption{\label{tab:Replica}\textbf{Quantitative Results on Replica.}}

\resizebox{\linewidth}{!}{
\begin{centering}
\begin{tabular}{ccccccccccc}
\hline 
\textbf{Cam} & \textbf{Method} & \textbf{Metric} & \textbf{office0} & \textbf{office1} & \textbf{office2} & \textbf{office3} & \textbf{office4} & \textbf{room0} & \textbf{room1} & \textbf{room2}\tabularnewline
\hline 
\multirow{10}{*}{\begin{turn}{90}
Monocular
\end{turn}} & \multirow{5}{*}{Photo-SLAM \cite{huang2024photo}} & ATE & $0.20\pm0.02$ & $2.95\pm6.23$ & $0.91\pm0.39$ & $0.11\pm0.01$ & $0.17\pm0.00$ & $0.15\pm0.00$ & $0.24\pm0.04$ & $0.10\pm0.02$\tabularnewline
 &  & FPS & $36.91\pm0.75$ & $36.41\pm0.66$ & $34.48\pm0.52$ & $34.60\pm0.36$ & $35.98\pm0.49$ & $34.40\pm0.29$ & $36.37\pm0.66$ & $33.32\pm0.28$\tabularnewline
 &  & IPF & $2.66\pm0.11$ & $2.31\pm0.05$ & $2.30\pm0.06$ & $2.29\pm0.05$ & $2.30\pm0.04$ & $2.03\pm0.02$ & $2.22\pm0.02$ & $2.30\pm0.08$\tabularnewline
 &  & PSNR & $35.02\pm0.45$ & $32.75\pm5.37$ & $31.19\pm0.65$ & $31.13\pm0.53$ & $32.94\pm0.18$ & $28.74\pm0.39$ & $30.56\pm0.38$ & $31.69\pm0.25$\tabularnewline
 &  & Points & \cellcolor{yellow!30}$78.40\text{k}\pm2.94\text{k}$ & $97.04\text{k}\pm31.16\text{k}$ & \cellcolor{yellow!30}$99.40\text{k}\pm1.67\text{k}$ & \cellcolor{yellow!30}$76.36\text{k}\pm3.19\text{k}$ & \cellcolor{yellow!30}$75.98\text{k}\pm3.39\text{k}$ & \cellcolor{yellow!30}$0.11\text{m}\pm6.27\text{k}$ & \cellcolor{yellow!30}$0.12\text{m}\pm5.56\text{k}$ & \cellcolor{orange!30}$81.10\text{k}\pm1.81\text{k}$\tabularnewline
\cline{2-11} \cline{3-11} \cline{4-11} \cline{5-11} \cline{6-11} \cline{7-11} \cline{8-11} \cline{9-11} \cline{10-11} \cline{11-11} 
 & \multirow{5}{*}{%
\begin{tabular}{c}
\textbf{Ours}\tabularnewline
\textbf{(Photo-SLAM)}\tabularnewline
\end{tabular}} & ATE & $0.22\pm0.06$ & $2.97\pm6.24$ & $1.53\pm1.37$ & $0.12\pm0.01$ & $0.17\pm0.01$ & $0.17\pm0.00$ & $0.52\pm0.48$ & $0.09\pm0.00$\tabularnewline
 &  & FPS & $36.65\pm0.46$ & $36.08\pm0.47$ & $33.90\pm0.28$ & $34.88\pm0.68$ & $35.96\pm0.54$ & $33.58\pm0.20$ & $36.65\pm0.29$ & $33.73\pm0.26$\tabularnewline
 &  & IPF & $8.10\pm0.21$ & $7.76\pm0.23$ & $8.05\pm0.13$ & $7.15\pm0.09$ & $7.35\pm0.13$ & $7.40\pm0.11$ & $7.33\pm0.24$ & $7.67\pm0.04$\tabularnewline
 &  & PSNR & $34.58\pm0.31$ & $34.97\pm4.96$ & \cellcolor{orange!30}$33.52\pm0.12$ & $33.26\pm0.08$ & $35.22\pm0.23$ & \cellcolor{yellow!30}$31.92\pm0.26$ & $31.99\pm1.15$ & $34.39\pm0.16$\tabularnewline
 &  & Points & \cellcolor{red!30}$\boldsymbol{38.32\text{k}\pm1.97\text{k}}$ & \cellcolor{red!30}$\boldsymbol{48.37\text{k}\pm11.77\text{k}}$ & \cellcolor{red!30}$\boldsymbol{64.07\text{k}\pm1.03\text{k}}$ & \cellcolor{red!30}$\boldsymbol{54.93\text{k}\pm0.91\text{k}}$ & \cellcolor{red!30}$\boldsymbol{53.67\text{k}\pm1.13\text{k}}$ & \cellcolor{red!30}$\boldsymbol{87.49\text{k}\pm2.99\text{k}}$ & \cellcolor{red!30}$\boldsymbol{73.44\text{k}\pm2.84\text{k}}$ & \cellcolor{red!30}$\boldsymbol{58.92\text{k}\pm1.30\text{k}}$\tabularnewline
\hline 
\multirow{20}{*}{\begin{turn}{90}
RGBD
\end{turn}} & \multirow{5}{*}{Photo-SLAM \cite{huang2024photo}} & ATE & $0.45\pm0.05$ & $0.35\pm0.04$ & $1.13\pm0.14$ & $0.37\pm0.02$ & $0.44\pm0.05$ & $0.30\pm0.02$ & $0.33\pm0.04$ & $0.18\pm0.00$\tabularnewline
 &  & FPS & $31.61\pm0.53$ & $31.96\pm0.32$ & $30.43\pm0.81$ & $29.33\pm0.52$ & $27.87\pm0.54$ & $27.49\pm0.52$ & $29.87\pm0.91$ & $27.37\pm0.52$\tabularnewline
 &  & IPF & $3.43\pm0.09$ & $3.04\pm0.12$ & $3.18\pm0.04$ & $3.28\pm0.04$ & $3.10\pm0.05$ & $3.17\pm0.05$ & $3.12\pm0.05$ & $3.20\pm0.05$\tabularnewline
 &  & PSNR & \cellcolor{yellow!30}$36.83\pm0.32$ & $36.79\pm0.29$ & $32.45\pm0.38$ & \cellcolor{yellow!30}$33.38\pm0.07$ & $35.13\pm0.39$ & $30.13\pm2.14$ & \cellcolor{orange!30}$33.80\pm0.36$ & $34.53\pm0.87$\tabularnewline
 &  & Points & $81.34\text{k}\pm2.95\text{k}$ & \cellcolor{yellow!30}$79.24\text{k}\pm1.71\text{k}$ & $0.12\text{m}\pm4.04\text{k}$ & $93.03\text{k}\pm3.79\text{k}$ & $0.12\text{m}\pm1.61\text{k}$ & $0.19\text{m}\pm2.70\text{k}$ & $0.16\text{m}\pm8.84\text{k}$ & $0.14\text{m}\pm2.09\text{k}$\tabularnewline
\cline{2-11} \cline{3-11} \cline{4-11} \cline{5-11} \cline{6-11} \cline{7-11} \cline{8-11} \cline{9-11} \cline{10-11} \cline{11-11} 
 & \multirow{5}{*}{%
\begin{tabular}{c}
\textbf{Ours}\tabularnewline
\textbf{(Photo-SLAM)}\tabularnewline
\end{tabular}} & ATE & $0.48\pm0.04$ & $0.38\pm0.06$ & $1.10\pm0.19$ & $0.38\pm0.02$ & $0.56\pm0.10$ & $0.31\pm0.01$ & $0.34\pm0.03$ & $0.18\pm0.00$\tabularnewline
 &  & FPS & $30.84\pm0.37$ & $31.49\pm0.31$ & $30.04\pm0.43$ & $28.76\pm0.58$ & $28.64\pm0.66$ & $27.81\pm0.62$ & $29.55\pm0.55$ & $26.87\pm0.31$\tabularnewline
 &  & IPF & $10.45\pm0.30$ & $9.90\pm0.26$ & $10.06\pm0.21$ & $10.40\pm0.40$ & $10.71\pm0.35$ & $9.95\pm0.66$ & $9.25\pm0.40$ & $9.97\pm0.06$\tabularnewline
 &  & PSNR & $35.54\pm0.28$ & \cellcolor{yellow!30}$37.74\pm0.41$ & \cellcolor{yellow!30}$33.40\pm0.29$ & \cellcolor{orange!30}$33.84\pm0.27$ & \cellcolor{yellow!30}$35.64\pm0.41$ & $29.38\pm3.70$ & \cellcolor{yellow!30}$34.30\pm0.64$ & \cellcolor{yellow!30}$36.54\pm0.19$\tabularnewline
 &  & Points & \cellcolor{orange!30}$39.74\text{k}\pm1.11\text{k}$ & \cellcolor{orange!30}$54.61\text{k}\pm2.58\text{k}$ & \cellcolor{orange!30}$79.29\text{k}\pm3.24\text{k}$ & \cellcolor{orange!30}$68.03\text{k}\pm2.06\text{k}$ & \cellcolor{orange!30}$75.58\text{k}\pm4.31\text{k}$ & \cellcolor{orange!30}$0.11\text{m}\pm3.74\text{k}$ & \cellcolor{orange!30}$0.10\text{m}\pm1.21\text{k}$ & \cellcolor{yellow!30}$0.10\text{m}\pm2.72\text{k}$\tabularnewline
\cline{2-11} \cline{3-11} \cline{4-11} \cline{5-11} \cline{6-11} \cline{7-11} \cline{8-11} \cline{9-11} \cline{10-11} \cline{11-11} 
 & \multirow{5}{*}{GS-ICP-SLAM\cite{ha2024rgbd}} & ATE & $0.19\pm0.00$ & $0.13\pm0.00$ & $0.18\pm0.00$ & $0.19\pm0.01$ & $0.22\pm0.01$ & $0.16\pm0.00$ & $0.16\pm0.00$ & $0.11\pm0.01$\tabularnewline
 &  & FPS & $30.00\pm0.00$ & $30.00\pm0.00$ & $30.00\pm0.00$ & $30.00\pm0.00$ & $30.00\pm0.00$ & $30.00\pm0.00$ & $30.00\pm0.00$ & $30.00\pm0.00$\tabularnewline
 &  & IPF & $2.88\pm0.00$ & $2.37\pm0.01$ & $2.88\pm0.01$ & $2.87\pm0.00$ & $2.91\pm0.01$ & $2.90\pm0.07$ & $2.84\pm0.07$ & $2.67\pm0.01$\tabularnewline
 &  & PSNR & \cellcolor{orange!30}$40.57\pm0.03$ & \cellcolor{orange!30}$40.96\pm0.11$ & $32.77\pm0.16$ & $31.60\pm0.07$ & \cellcolor{orange!30}$38.84\pm0.04$ & \cellcolor{orange!30}$35.54\pm0.06$ & \cellcolor{orange!30}$37.81\pm0.06$ & \cellcolor{orange!30}$38.54\pm0.05$\tabularnewline
 &  & Points & $1.57\text{m}\pm0.85\text{k}$ & $1.57\text{m}\pm7.30\text{k}$ & $1.54\text{m}\pm2.51\text{k}$ & $1.55\text{m}\pm9.54\text{k}$ & $1.60\text{m}\pm10.33\text{k}$ & $1.55\text{m}\pm2.86\text{k}$ & $1.55\text{m}\pm0.70\text{k}$ & $1.54\text{m}\pm3.78\text{k}$\tabularnewline
\cline{2-11} \cline{3-11} \cline{4-11} \cline{5-11} \cline{6-11} \cline{7-11} \cline{8-11} \cline{9-11} \cline{10-11} \cline{11-11} 
 & \multirow{5}{*}{%
\begin{tabular}{c}
\textbf{Ours}\tabularnewline
\textbf{(GS-ICP-SLAM)}\tabularnewline
\end{tabular}} & ATE & $0.25\pm0.14$ & $0.12\pm0.00$ & $0.28\pm0.10$ & $0.19\pm0.02$ & $0.24\pm0.01$ & $0.16\pm0.00$ & $0.16\pm0.00$ & $0.11\pm0.00$\tabularnewline
 &  & FPS & $30.00\pm0.00$ & $30.00\pm0.00$ & $30.00\pm0.00$ & $30.00\pm0.00$ & $30.00\pm0.00$ & $30.00\pm0.00$ & $30.00\pm0.00$ & $30.00\pm0.00$\tabularnewline
 &  & IPF & $12.10\pm0.07$ & $12.36\pm0.05$ & $10.42\pm0.08$ & $10.97\pm0.04$ & $11.47\pm0.05$ & $11.03\pm0.03$ & $11.68\pm0.08$ & $10.46\pm0.05$\tabularnewline
 &  & PSNR & \cellcolor{red!30}$\boldsymbol{42.60\pm0.05}$ & \cellcolor{red!30}$\boldsymbol{42.33\pm0.03}$ & \cellcolor{red!30}$\boldsymbol{36.95\pm0.42}$ & \cellcolor{red!30}$\boldsymbol{36.87\pm0.03}$ & \cellcolor{red!30}$\boldsymbol{39.77\pm0.08}$ & \cellcolor{red!30}$\boldsymbol{36.46\pm0.10}$ & \cellcolor{red!30}$\boldsymbol{39.19\pm0.05}$ & \cellcolor{red!30}$\boldsymbol{39.38\pm0.23}$\tabularnewline
 &  & Points & $0.76\text{m}\pm7.29\text{k}$ & $0.67\text{m}\pm5.51\text{k}$ & $0.79\text{m}\pm17.16\text{k}$ & $0.74\text{m}\pm12.55\text{k}$ & $0.70\text{m}\pm16.92\text{k}$ & $0.72\text{m}\pm16.73\text{k}$ & $0.72\text{m}\pm16.60\text{k}$ & $0.70\text{m}\pm13.53\text{k}$\tabularnewline
\hline 
\end{tabular}
\par\end{centering}
}

\vspace{-2mm}
\end{table*}

\begin{table}
\caption{\label{tab:TUM-RGBD}\textbf{Quantitative Results on TUM-RGBD.}}

\resizebox{\linewidth}{!}{
\begin{centering}
\begin{tabular}{cccccc}
\hline 
\textbf{Cam} & \textbf{Method} & \textbf{Metric} & \textbf{fr1/desk} & \textbf{fr2/xyz} & \textbf{fr3/office}\tabularnewline
\hline 
\multirow{15}{*}{\begin{turn}{90}
Monocular
\end{turn}} & \multirow{5}{*}{MonoGS \cite{matsuki2024gaussian}} & ATE & $4.93\pm0.16$ & $4.66\pm0.13$ & $3.35\pm0.45$\tabularnewline
 &  & FPS & $1.87\pm0.05$ & $3.37\pm0.06$ & $2.26\pm0.01$\tabularnewline
 &  & IPF & $84.07\pm0.25$ & $51.64\pm0.26$ & $60.5\pm0.43$\tabularnewline
 &  & PSNR & $17.65\pm0.40$ & $15.56\pm0.02$ & $19.35\pm0.31$\tabularnewline
 &  & Points & \cellcolor{red!30}$\boldsymbol{26.64\text{k}\pm1.58\text{k}}$ & \cellcolor{orange!30}$43.59\text{k}\pm2.09\text{k}$ & \cellcolor{red!30}$\boldsymbol{35.24\text{k}\pm3.24\text{k}}$\tabularnewline
\cline{2-6} \cline{3-6} \cline{4-6} \cline{5-6} \cline{6-6} 
 & \multirow{5}{*}{Photo-SLAM \cite{huang2024photo}} & ATE & $1.55\pm0.06$ & $0.63\pm0.18$ & $1.10\pm0.70$\tabularnewline
 &  & FPS & $25.18\pm0.30$ & $25.83\pm0.12$ & $24.74\pm0.25$\tabularnewline
 &  & IPF & $7.08\pm0.08$ & $6.66\pm0.08$ & $7.77\pm0.20$\tabularnewline
 &  & PSNR & $19.69\pm0.04$ & $20.19\pm0.52$ & $18.32\pm1.36$\tabularnewline
 &  & Points & $40.00\text{k}\pm0.79\text{k}$ & $0.10\text{m}\pm7.50\text{k}$ & $81.16\text{k}\pm3.44\text{k}$\tabularnewline
\cline{2-6} \cline{3-6} \cline{4-6} \cline{5-6} \cline{6-6} 
 & \multirow{5}{*}{%
\begin{tabular}{c}
\textbf{Ours}\tabularnewline
\textbf{(Photo-SLAM)}\tabularnewline
\end{tabular}} & ATE & $1.55\pm0.06$ & $0.70\pm0.08$ & $0.57\pm0.33$\tabularnewline
 &  & FPS & $24.95\pm0.46$ & $26.16\pm0.12$ & $25.03\pm0.11$\tabularnewline
 &  & IPF & $17.88\pm0.02$ & $14.41\pm0.26$ & $16.06\pm0.32$\tabularnewline
 &  & PSNR & $20.51\pm0.08$ & $21.54\pm0.85$ & $19.38\pm1.47$\tabularnewline
 &  & Points & \cellcolor{yellow!30}$38.65\text{k}\pm1.82\text{k}$ & $66.51\text{k}\pm1.71\text{k}$ & \cellcolor{orange!30}$51.71\text{k}\pm3.46\text{k}$\tabularnewline
\hline 
\multirow{40}{*}{\begin{turn}{90}
RGBD
\end{turn}} & \multirow{5}{*}{Loopy-SLAM \cite{liso2024loopy}} & ATE & $3.93\pm1.13$ & $1.43\pm0.16$ & $4.65\pm1.63$\tabularnewline
 &  & FPS & $0.23\pm0.00$ & $0.21\pm0.00$ & $0.20\pm0.00$\tabularnewline
 &  & IPF & - & - & -\tabularnewline
 &  & PSNR & $13.66\pm0.12$ & $17.95\pm0.41$ & $17.43\pm0.15$\tabularnewline
 &  & Points & - & - & -\tabularnewline
\cline{2-6} \cline{3-6} \cline{4-6} \cline{5-6} \cline{6-6} 
 & \multirow{5}{*}{SplaTAM \cite{keetha2024splatam}} & ATE & $2.51\pm0.01$ & $0.50\pm0.00$ & $4.52\pm0.21$\tabularnewline
 &  & FPS & $0.27\pm0.01$ & $0.03\pm0.02$ & $0.25\pm0.00$\tabularnewline
 &  & IPF & $460.32\pm0.00$ & $460.88\pm0.00$ & $460.84\pm0.00$\tabularnewline
 &  & PSNR & \cellcolor{orange!30}$21.03\pm0.10$ & \cellcolor{orange!30}$23.19\pm0.13$ & $20.10\pm0.05$\tabularnewline
 &  & Points & $0.96\text{m}\pm3.96\text{k}$ & $6.36\text{m}\pm81.37\text{k}$ & $0.79\text{m}\pm5.89\text{k}$\tabularnewline
\cline{2-6} \cline{3-6} \cline{4-6} \cline{5-6} \cline{6-6} 
 & \multirow{5}{*}{Gaussian-SLAM \cite{yugay2023gaussian}} & ATE & $2.74\pm0.11$ & $0.96\pm0.44$ & $8.42\pm1.19$\tabularnewline
 &  & FPS & $0.57\pm0.06$ & $0.48\pm0.03$ & $0.59\pm0.02$\tabularnewline
 &  & IPF & $309.37\pm4.29$ & $308.44\pm0.04$ & $310.66\pm0.11$\tabularnewline
 &  & PSNR & \cellcolor{red!30}$\boldsymbol{23.71\pm0.10}$ & \cellcolor{red!30}$\boldsymbol{23.95\pm0.39}$ & \cellcolor{red!30}$\boldsymbol{25.80\pm0.09}$\tabularnewline
 &  & Points & $0.76\text{m}\pm12.12\text{k}$ & $0.69\text{m}\pm26.07\text{k}$ & $1.47\text{m}\pm6.75\text{k}$\tabularnewline
\cline{2-6} \cline{3-6} \cline{4-6} \cline{5-6} \cline{6-6} 
 & \multirow{5}{*}{MonoGS \cite{matsuki2024gaussian}} & ATE & $1.84\pm0.09$ & $1.71\pm0.08$ & $1.74\pm0.10$\tabularnewline
 &  & FPS & $2.18\pm0.02$ & $3.23\pm0.07$ & $2.48\pm0.03$\tabularnewline
 &  & IPF & $77.77\pm0.06$ & $51.23\pm0.18$ & $63.20\pm0.06$\tabularnewline
 &  & PSNR & $19.00\pm0.09$ & $15.81\pm0.03$ & $19.11\pm0.25$\tabularnewline
 &  & Points & $43.01\text{k}\pm1.95\text{k}$ & \cellcolor{red!30}$\boldsymbol{37.20\text{k}\pm4.78\text{k}}$ & \cellcolor{yellow!30}$52.67\text{k}\pm2.00\text{k}$\tabularnewline
\cline{2-6} \cline{3-6} \cline{4-6} \cline{5-6} \cline{6-6} 
 & \multirow{5}{*}{Photo-SLAM \cite{huang2024photo}} & ATE & $1.49\pm0.03$ & $0.32\pm0.02$ & $1.17\pm0.34$\tabularnewline
 &  & FPS & $23.45\pm0.18$ & $23.44\pm0.01$ & $22.63\pm0.22$\tabularnewline
 &  & IPF & $8.88\pm0.14$ & $7.68\pm0.28$ & $8.54\pm0.26$\tabularnewline
 &  & PSNR & $19.98\pm0.03$ & $21.92\pm0.42$ & \cellcolor{yellow!30}$22.18\pm1.20$\tabularnewline
 &  & Points & $45.64\text{k}\pm1.18\text{k}$ & $68.68\text{k}\pm10.00\text{k}$ & $67.69\text{k}\pm1.75\text{k}$\tabularnewline
\cline{2-6} \cline{3-6} \cline{4-6} \cline{5-6} \cline{6-6} 
 & \multirow{5}{*}{%
\begin{tabular}{c}
\textbf{Ours}\tabularnewline
\textbf{(Photo-SLAM)}\tabularnewline
\end{tabular}} & ATE & $1.52\pm0.03$ & $0.30\pm0.01$ & $0.90\pm0.03$\tabularnewline
 &  & FPS & $23.06\pm0.22$ & $23.36\pm0.07$ & $22.78\pm0.10$\tabularnewline
 &  & IPF & $20.60\pm0.46$ & $18.05\pm0.31$ & $17.66\pm0.32$\tabularnewline
 &  & PSNR & \cellcolor{yellow!30}$20.54\pm0.06$ & \cellcolor{yellow!30}$22.75\pm0.22$ & \cellcolor{orange!30}$22.95\pm0.79$\tabularnewline
 &  & Points & \cellcolor{orange!30}$38.65\text{k}\pm0.76\text{k}$ & \cellcolor{yellow!30}$49.80\text{k}\pm2.63\text{k}$ & $71.33\text{k}\pm6.79\text{k}$\tabularnewline
\cline{2-6} \cline{3-6} \cline{4-6} \cline{5-6} \cline{6-6} 
 & \multirow{5}{*}{GS-ICP-SLAM \cite{ha2024rgbd}} & ATE & $3.26\pm0.28$ & $2.26\pm0.04$ & $3.07\pm0.41$\tabularnewline
 &  & FPS & $30.00\pm0.00$ & $30.00\pm0.00$ & $30.00\pm0.00$\tabularnewline
 &  & IPF & $6.10\pm0.05$ & $3.69\pm0.05$ & $3.96\pm0.08$\tabularnewline
 &  & PSNR & $15.62\pm0.07$ & $18.43\pm0.19$ & $19.20\pm0.05$\tabularnewline
 &  & Points & $0.53\text{m}\pm6.82\text{k}$ & $1.91\text{m}\pm11.37\text{k}$ & $2.09\text{m}\pm21.04\text{k}$\tabularnewline
\cline{2-6} \cline{3-6} \cline{4-6} \cline{5-6} \cline{6-6} 
 & \multirow{5}{*}{%
\begin{tabular}{c}
\textbf{Ours}\tabularnewline
\textbf{(GS-ICP-SLAM)}\tabularnewline
\end{tabular}} & ATE & $3.92\pm0.71$ & $2.44\pm0.06$ & $4.11\pm1.28$\tabularnewline
 &  & FPS & $30.00\pm0.00$ & $30.00\pm0.00$ & $30.00\pm0.00$\tabularnewline
 &  & IPF & $20.02\pm0.10$ & $18.43\pm0.15$ & $12.17\pm0.13$\tabularnewline
 &  & PSNR & $17.54\pm0.07$ & $21.35\pm0.20$ & $20.84\pm0.06$\tabularnewline
 &  & Points & $0.18\text{m}\pm3.65\text{k}$ & $0.13\text{m}\pm12.32\text{k}$ & $0.34\text{m}\pm19.24\text{k}$\tabularnewline
\hline 
\end{tabular}
\par\end{centering}
}

\vspace{-2mm}
\end{table}

\begin{table}
\caption{\textbf{\textcolor{red}{\label{tab:VECtor}}}\textbf{Quantitative
Results on VECtor.}}

\resizebox{\linewidth}{!}{
\begin{centering}
\begin{tabular}{cccccc}
\hline 
\textbf{Cam} & \textbf{Method} & \textbf{Metric} & \textbf{corner-slow} & \textbf{robot-normal} & \textbf{corridors-dolly}\tabularnewline
\hline 
\multirow{10}{*}{\begin{turn}{90}
Monocular
\end{turn}} & \multirow{5}{*}{Photo-SLAM \cite{huang2024photo}} & ATE & $0.66\pm0.01$ & $2.20\pm1.66$ & $9.56\pm6.08$\tabularnewline
 &  & FPS & $23.27\pm0.21$ & $21.90\pm0.32$ & $20.18\pm0.26$\tabularnewline
 &  & IPF & $3.11\pm0.03$ & $3.37\pm0.17$ & $3.11\pm0.03$\tabularnewline
 &  & PSNR & \cellcolor{orange!30}$24.63\pm0.05$ & \cellcolor{orange!30}$19.58\pm0.18$ & \cellcolor{yellow!30}$15.31\pm0.69$\tabularnewline
 &  & Points & $0.12\text{m}\pm17.02\text{k}$ & $0.16\text{m}\pm72.38\text{k}$ & $0.38\text{m}\pm3.99\text{k}$\tabularnewline
\cline{2-6} \cline{3-6} \cline{4-6} \cline{5-6} \cline{6-6} 
 & \multirow{5}{*}{%
\begin{tabular}{c}
\textbf{Ours}\tabularnewline
\textbf{(Photo-SLAM)}\tabularnewline
\end{tabular}} & ATE & $0.68\pm0.02$ & $2.35\pm1.17$ & $10.06\pm6.20$\tabularnewline
 &  & FPS & $21.56\pm0.33$ & $18.30\pm1.20$ & $18.00\pm0.26$\tabularnewline
 &  & IPF & $7.69\pm0.17$ & $10.78\pm0.68$ & $11.11\pm0.18$\tabularnewline
 &  & PSNR & \cellcolor{red!30}$\boldsymbol{25.37\pm0.12}$ & \cellcolor{red!30}$\boldsymbol{22.16\pm1.46}$ & \cellcolor{red!30}$\boldsymbol{23.02\pm5.67}$\tabularnewline
 &  & Points & \cellcolor{orange!30}$7.31\text{k}\pm0.25\text{k}$ & \cellcolor{orange!30}$8.24\text{k}\pm2.06\text{k}$ & \cellcolor{orange!30}$36.96\text{k}\pm1.59\text{k}$\tabularnewline
\hline 
\multirow{10}{*}{\begin{turn}{90}
Stereo
\end{turn}} & \multirow{5}{*}{Photo-SLAM \cite{huang2024photo}} & ATE & $1.15\pm0.00$ & $1.52\pm0.00$ & $11.91\pm0.04$\tabularnewline
 &  & FPS & $20.43\pm0.32$ & $17.77\pm0.31$ & $19.31\pm0.01$\tabularnewline
 &  & IPF & $1.68\pm0.08$ & $2.58\pm0.04$ & $2.76\pm0.02$\tabularnewline
 &  & PSNR & $19.34\pm0.02$ & $16.59\pm0.01$ & $14.51\pm0.34$\tabularnewline
 &  & Points & \cellcolor{yellow!30}$38.98\text{k}\pm4.29\text{k}$ & \cellcolor{yellow!30}$47.36\text{k}\pm0.64\text{k}$ & \cellcolor{yellow!30}$0.24\text{m}\pm2.92\text{k}$\tabularnewline
\cline{2-6} \cline{3-6} \cline{4-6} \cline{5-6} \cline{6-6} 
 & \multirow{5}{*}{%
\begin{tabular}{c}
\textbf{Ours}\tabularnewline
\textbf{(Photo-SLAM)}\tabularnewline
\end{tabular}} & ATE & $1.15\pm0.00$ & $1.52\pm0.00$ & $11.51\pm0.23$\tabularnewline
 &  & FPS & $20.75\pm0.37$ & $14.64\pm0.23$ & $16.64\pm0.83$\tabularnewline
 &  & IPF & $9.23\pm0.02$ & $12.24\pm0.20$ & $11.21\pm0.16$\tabularnewline
 &  & PSNR & \cellcolor{yellow!30}$19.56\pm0.04$ & \cellcolor{yellow!30}$16.77\pm0.05$ & \cellcolor{orange!30}$19.34\pm0.06$\tabularnewline
 &  & Points & \cellcolor{red!30}$\boldsymbol{6.45\text{k}\pm0.20\text{k}}$ & \cellcolor{red!30}$\boldsymbol{7.68\text{k}\pm0.24\text{k}}$ & \cellcolor{red!30}$\boldsymbol{30.81\text{k}\pm2.21\text{k}}$\tabularnewline
\hline 
\end{tabular}
\par\end{centering}
}

\vspace{-6mm}
\end{table}

\noindent where $r_{k+1}^{0}$ is the initial optimization iteration
count assigned to the new keyframe, and $l_{k+1}$ is its initial
optimization loss value. We then select a keyframe $v'$ randomly
from the subset of keyframes with remaining iterations, defined as
$\left\{ v_{i}\vert r_{i}>0,\forall r_{i}\in\mathcal{R}_{k}\right\} $,
to train the 3D Gaussians Map $\mathcal{G}$. Post-optimization, we
decrement the optimization iteration count for the selected keyframe
by one, adjusting $r'$ to $r'-1$, and also update the corresponding
optimization loss value $l'$.

When $\left\{ v_{i}\vert r_{i}>0,\forall r_{i}\in\mathcal{R}_{k}\right\} $
is empty, we update $\mathcal{R}_{k}$ based on $\mathcal{L}_{k}$
as follows:

\vspace{-2mm}

\begin{equation}
r_{i}=\begin{cases}
1 & l_{i}\notin\overset{d_{k}}{\prod}(\mathcal{L}_{k}),\\
2 & l_{i}\in\overset{d_{k}}{\prod}(\mathcal{L}_{k}),
\end{cases}
\end{equation}

\noindent where $\overset{d_{k}}{\prod}(\cdot)$ donates top $d_{k}$
largest elements, $d_{k}=\max(1,\frac{k}{d})$, and $d$ is a hyperparameter.
This method prioritizes keyframes with higher optimization loss values
for the photorealistic rendering module, effectively tackling the
long-tail optimization as demonstrated in \Figref{adaptive-opt}.

\subsubsection{\label{subsec:Opacity-Regularization}Opacity Regularization}

In the typical application of 3DGS, the rendered loss $\mathcal{L}_{rendered}$
is utilized to refine the 3D Gaussian primitives \cite{kerbl20233d}.
To efficiently manage memory usage and model size, we have devised
a strategy that encourages the elimination of Gaussians in areas where
they do not contribute to the rendering process. Since the presence
of a Gaussian is primarily indicated by its opacity $o$, we impose
a regularization term $\mathcal{L}_{o}$ on this attribute. The complete
formulation of our optimization loss $\mathcal{L}$ is as follows:

\vspace{-3mm}

\begin{equation}
\mathcal{L}_{rendered}=(1-\lambda_{ssim})\mathcal{L}_{1}+\lambda_{ssim}\mathcal{L}_{ssim},
\end{equation}

\vspace{-3mm}

\begin{equation}
\mathcal{L}_{o}=\frac{1}{N}\sum_{i}\vert o_{i}\vert,
\end{equation}

\vspace{-3mm}

\begin{equation}
\mathcal{L}=\mathcal{L}_{rendered}+\lambda_{o}\mathcal{L}_{o},
\end{equation}

\noindent where $\lambda_{ssim}$ is the weighting factor, $\lambda_{o}$
is the regularization coefficient, and $N$ denotes the total count
of Gaussian primitives.

\section{Experiments}

In this section, we present a comparative analysis of CaRtGS against
state-of-the-art GS-SLAM systems \cite{matsuki2024gaussian,keetha2024splatam,yugay2023gaussian,huang2024photo,ha2024rgbd}
and Loopy-SLAM \cite{liso2024loopy}, a state-of-the-art NeRF-based
SLAM system. This evaluation spans multiple scenarios, including those
captured using monocular, RGB-D, and stereo cameras. Furthermore,
we perform an ablation study to substantiate the efficacy of the novel
techniques introduced in our approach.

\begin{figure*}
\begin{centering}
\includegraphics[width=1\linewidth]{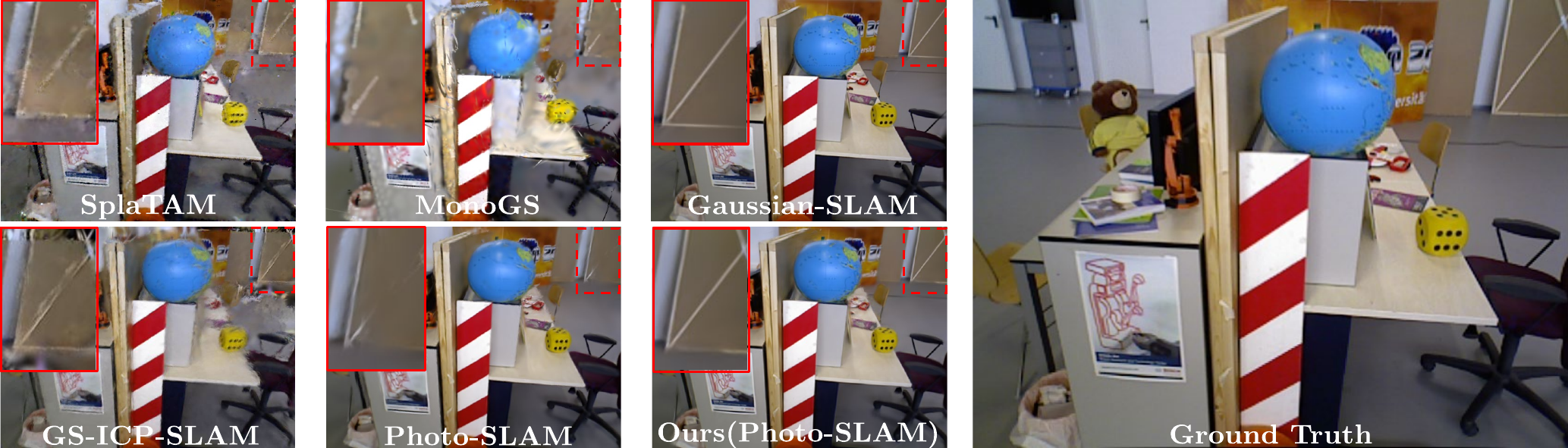}
\par\end{centering}
\caption{{\small{}\label{fig:TUM-RGBD}}\textbf{Qualitative results on TUM-RGBD
with RGBD Camera.}\textbf{\small{} }{\small{}Qualitative assessments
demonstrate that our approach significantly improves rendering quality
and effectively mitigates visual artifacts. Furthermore, our method
achieves precise localization accuracy. In contrast, Gaussian-SLAM
exhibits substantial drift, as indicated by the red dashed line.}}

\vspace{-6mm}
\end{figure*}

\subsection{Setup}

\textbf{Dataset.} We conducted evaluations on three distinct camera
systems: monocular, RGB-D, and stereo. These assessments were carried
out on three renowned datasets: Replica \cite{straub2019replica},
TUM-RGBD \cite{sturm2012benchmark}, and VECtor \cite{gao2022vector}.
Replica \cite{straub2019replica} is a high-quality reconstruction
dataset at room and building scale, including high-resolution high-dynamic-range
(HDR) textures. TUM-RGBD \cite{sturm2012benchmark} is a well-known
RGB-D dataset that contains color and depth images captured by a Microsoft
Kinect sensor, along with the ground-truth trajectory obtained from
a high-accuracy motion-capture system. VECtor \cite{gao2022vector}
is a SLAM benchmark dataset that covers the full spectrum of motion
dynamics, environmental complexities, and illumination conditions.
To ensure data consistency, we employed a soft time synchronization
to align the sensor data and ground truth with a precision of $\Delta t=0.08s$.

\textbf{Implementation Detail.} All experimental evaluations were
conducted on a desktop with an \textit{Nvidia RTX 4090} \textit{GPU},
an \textit{AMD Ryzen 9 7950X CPU}, and \textit{128 GB RAM}. We retained
most of the original hyperparameters from the 3DGS \cite{kerbl20233d}.
However, we densify every 500 iterations with a positional gradients
threshold $\tau_{p}=0.001$ and remove the transparent Gaussians with
a threshold $\epsilon_{\alpha}=0.02$. By default, we set $d=4$ and
$\lambda_{o}=0.001$. On Replica, we use $r_{k+1}^{0}=8$, whileas
$r_{k+1}^{0}=2$ on TUM-RGBD and VECtor.

\textbf{Evaluation.} We performed all experiments $5$ times to ensure
statistical robustness and rendered original-resolution images for
each estimated camera pose. To measure performance, we utilized the
evo toolkit\footnote{\url{https://github.com/MichaelGrupp/evo}} and
the torchmetrics toolkit\footnote{\url{https://github.com/Lightning-AI/torchmetrics}}.
We recorded various performance indicators, including Absolute Trajectory
Error (ATE) to assess the accuracy of localization, Peak Signal-to-Noise
Ratio (PSNR) to assess the quality of the photorealistic renderings,
and the number of 3D Gaussian points to assess the model size. To
assess the sufficiency of the Gaussian primitives' optimization, we
introduced a metric known as Iterations Per Frame (IPF), defined as
the ratio of total iterations to the total number of frames ($\text{IPF}=\frac{\text{Iterations}}{\text{Frames}}$).\textcolor{red}{{}
}All performance indicators are reported in the format of mean $\pm$
standard deviation.

\subsection{Results}

The quantitative comparison presented in \Tabref{Replica}, \Tabref{TUM-RGBD},
and \Tabref{VECtor} illustrates the performance of various methods.
The best resutls of the PSNR and the count of Gaussian primitives
are distinctively highlighted as \colorbox{red!30}{$\boldsymbol{1^{\textbf{st}}}$},
\colorbox{orange!30}{$2^{\text{nd}}$}, and \colorbox{yellow!30}{$3^{\text{rd}}$}.
In summary, our approach consistently delivers superior rendering
performance, utilizing a reduced number of Gaussian primitives, while
adhering to real-time constraints of over 22 frames per second. Specifically,
on the Replica dataset \cite{straub2019replica} with monocular camera,
compared with Photo-SLAM \cite{huang2024photo}, and under similar
localization accuracy, our approach significantly improves the average
PSNR by more than $2\,\text{dB}$ and halves the number of Gaussian
primitives. As shown in \Tabref{Replica} and \Tabref{TUM-RGBD},
our method can be easily integrated into Photo-SLAM \cite{huang2024photo}
and GS-ICP-SLAM \cite{ha2024rgbd}. In \Tabref{TUM-RGBD}, our approach
achieves high rendering quality using a comparable number of Gaussian
primitives to MonoGS \cite{matsuki2024gaussian}. In \Tabref{VECtor},
we present the results on VECtor \cite{gao2022vector}, specifically
using a monocular camera. Our method improves the average PSNR by
more than $3\,\text{dB}$ with only one-tenth of the Gaussian primitives.
Furthermore, the qualitative results depicted in \Figref{TUM-RGBD}
corroborate that our approach achieves high-fidelity rendering.

\begin{figure}
\begin{centering}
\includegraphics[width=0.9\linewidth]{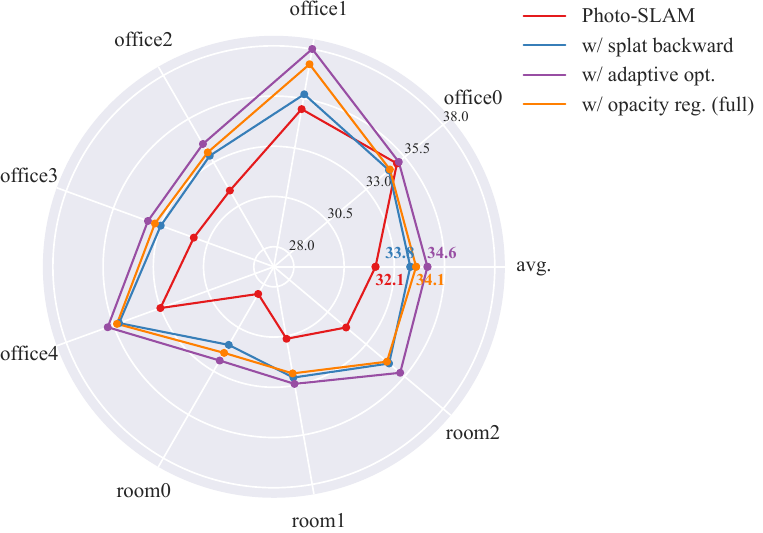}
\par\end{centering}
\caption{{\small{}\label{fig:Ablation-Study}}\textbf{\small{}The Radar Chart
of Ablation Study. }{\small{}Radial axis presents the PSNR.}}

\vspace{-4mm}
\end{figure}

Figure \ref{fig:Ablation-Study} depicts our ablation studies on the
monocular Replica dataset \cite{straub2019replica}, rigorously validating
our design choices and highlighting their contributions to system
performance. Key findings include:

\textbf{Splat-wise backpropagation enhances the rendering quality
by refining the iterative process efficiently.} The integration of
splat-wise backpropagation has significantly improved average total
iterations from $4.6\text{k}$ to $15.4\text{k}$ and average PSNR
from $32.1$~dB to $33.8$~dB.

\textbf{Adaptive optimization strategically allocates computational
resources to enhance rendering quality.} Integrating splat-wise backpropagation
with adaptive optimization has continuously boosted average PSNR from
$33.8$~dB to $34.6$~dB. Furthermore, as illustrated in \Figref{adaptive-opt},
this approach equitably distributes computational resources across
keyframes, efficiently addressing long-tail optimization challenges.

\begin{figure}
\begin{centering}
\includegraphics[width=1\linewidth]{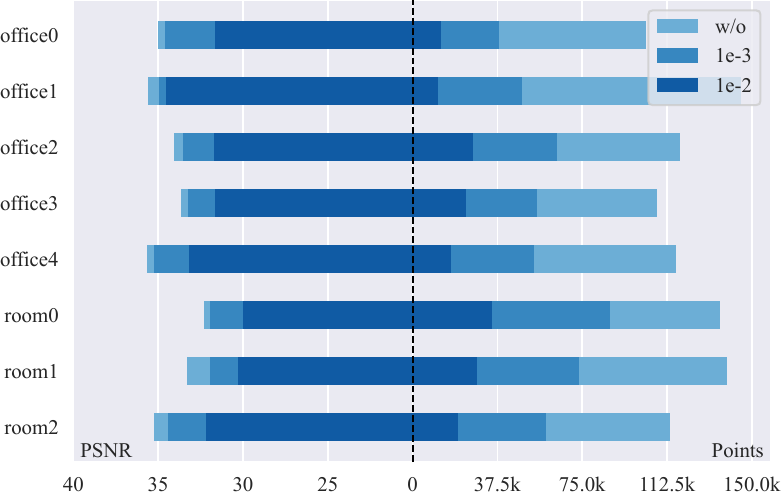}
\par\end{centering}
\caption{{\small{}\label{fig:opacity-reg}}\textbf{The Effect of Opacity Regularization.}\textbf{\small{}
}{\small{}The left side illustrates the value of PSNR. The right side
depicts the count of Gaussian points.}}

\vspace{-6mm}
\end{figure}

\textbf{Opacity regularization is instrumental in reducing the model
size without compromising the superior rendering quality.} Our opacity
regularization technique, as shown in \Figref{opacity-reg}, can halve
the model size with a regularization coefficient of $\lambda_{o}=0.001$,
with minimal PSNR performance loss.\textcolor{red}{{} }Increasing the
coefficient to $0.01$ further reduces less critical Gaussian primitives,
which results in a more efficient model at the expense of some rendering
quality.

\section{Limitations and Feature Work}

CaRtGS is an adaptive optimization technology that leverages 3D Gaussian
models for high-quality rendering and environmental reconstruction
in real-time GS-SLAM systems. Despite its potential, several limitations
and challenges are identified below, structured into categories for
clarity:
\begin{enumerate}
\item \textbf{Dynamic Environment Challenges.} CaRtGS assumes static environments,
limiting real-world use and causing tracking failures with dynamic
objects.
\item \textbf{Localization Robustness.} CaRtGS focuses on improving the
rendering quality of GS-SLAM. However, localization accuracy affects
rendering quality, especially in some degeneracy scenarios. Therefore,
a robustness localization module is essential for GS-SLAM.
\item \textbf{Geometry Accuracy.} Effective geometry mapping is vital in
GS-SLAM. As shown in \Tabref{VECtor}, the stereo model's inferior
rendering quality stems from the stereo camera's suboptimal geometry
mapping.
\end{enumerate}
Looking forward, we envision further improvements by integrating advanced
machine learning models to predict and handle dynamic objects.

\section{Conclusion}

In this work, we introduced CaRtGS, a novel framework that integrates
computational alignment with Gaussian Splatting SLAM to achieve real-time
photorealistic dense rendering. Our key contribution lies in the development
of an adaptive computational alignment strategy that optimizes the
rendering process by addressing the computational misalignment inherent
in GS-SLAM systems. Through fast splat-wise backpropagation, adaptive
optimization, and opacity regularization, we significantly enhanced
the rendering quality and computational efficiency of the SLAM process.

\bibliographystyle{IEEEtran}
\bibliography{bibtex/ref}

\end{document}